\documentclass[journal, onecolumn]{IEEEtran}

\usepackage[utf8]{inputenc}
\usepackage{amsmath}

\usepackage[T1]{fontenc} 

\usepackage{hyperref} 

\usepackage{mathtools}
\usepackage{graphicx}
\usepackage{tabularx}
\usepackage{subfigure}
\usepackage[table]{xcolor}
\graphicspath{{images/}}
\usepackage{dblfloatfix}
\usepackage{fixltx2e}
\usepackage{multirow}
\usepackage[justification=centering]{caption}

\usepackage{tabularx} 
\usepackage{array} 
\usepackage{booktabs} 

\usepackage{multirow}
\usepackage[table]{xcolor}

\usepackage{amsmath} 
\usepackage{amssymb} 
\usepackage{mathrsfs} 
\usepackage{amsthm} 
\usepackage{mathtools} 

\usepackage{amssymb}
\usepackage[linesnumbered,ruled,vlined]{algorithm2e}
\usepackage{xcolor}
\usepackage{multirow}
\usepackage{float}
\usepackage{hyperref}
\usepackage{comment}
\usepackage[nameinlink]{cleveref} 

\begin{document}
\title{Overcoming Catastrophic Forgetting in Tabular Data Classification: A Pseudorehearsal-based approach}
\author{Pablo García-Santaclara, Bruno Fernández-Castro, Rebeca P. Díaz-Redondo

\thanks{Pablo García-Santaclara (pgarcia@alumnos.uvigo.es) and Rebeca P. Díaz-Redondo (rebeca@det.uvigo.es) are with atlanTTic - I\&C Lab - Universidade de Vigo, Universidade de Vigo; Vigo, 36310, Spain.}

\thanks{Pablo García-Santaclara (pgsantaclara@gradiant.org) and Bruno Fernández-Castro (bfernandez@gradiant.org) are with Centro Tecnolóxico de Telecomunicacións de Galicia (GRADIANT),  Carretera do Vilar, 56-58, 36214, Vigo, Spain}
}

\maketitle

\begin{abstract}

\noindent Continual learning (CL) poses the important challenge of adapting to evolving data distributions without forgetting previously acquired knowledge while consolidating new knowledge. In this paper, we introduce a new methodology, coined as Tabular-data Rehearsal-based Incremental Lifelong Learning framework (TRIL3), designed to address the phenomenon of catastrophic forgetting in tabular data classification problems. TRIL3 uses the prototype-based incremental generative model XuILVQ to generate synthetic data to preserve old knowledge and the DNDF algorithm, which was modified to run in an incremental way, to learn classification tasks for tabular data, without storing old samples. After different tests to obtain the adequate percentage of synthetic data and to compare TRIL3 with other CL available proposals, we can conclude that the performance of TRIL3 outstands other options in the literature using only $50\%$ of synthetic data.

\end{abstract}


\section{Introduction} 

Continual learning (CL)~\cite{de2021continual, wang2024comprehensive}, also known as lifelong learning~\cite{parisi2019continual}, is an artificial intelligence approach that focuses on the ability of models to adapt and improve over time as they incrementally learn while processing dynamic data-streams. The underlying philosophy is using batches of data, a batch may be even just one sample, taken from a data-stream to train the system: each batch is used only once. This means that it is not possible to access previously processed data and, therefore, this entails a radical change compared to the classical pipeline of training, validating, and testing in ML. 

Therefore, CL is highly recommendable when facing scenarios where the model needs to adapt quickly to new data or when the model needs to be personalized. However, there is an important challenge in CL due to its nature: since models easily adapt to new knowledge, they tend to forget past knowledge. This effect, known as catastrophic forgetting~\cite{french1999catastrophic, kirkpatrick2017overcoming}, entails models to reduce their performance when acquiring new knowledge, which impacts their usefulness. This is especially severe in class-incremental learning, when it is expected the model is able to differentiate among a set of classes. When new classes appear, the model learns to identify them but this new knowledge alters the decision boundaries of the previous classes, causing the model to deteriorate.




Although there are multiple approaches to mitigate catastrophic forgetting, rehearsal methods~\cite{robins1995catastrophic} are one of the most widely used. They are based on combining new samples with representative samples of past knowledge, which can be obtained by keeping old samples or by employing generative models to map statistical distributions of the training data. Both, (i) the old or generated synthetic samples and (ii) new and unseen data are combined to train CL models. Despite of its advantages, rehearsal methods do not solve all drawbacks of catastrophic forgetting~\cite{verwimp2021rehearsal}. On the one hand, these methods are usually implemented through deep learning networks that reliably reproduce past data distributions. However, they do not work in an incremental way, so they are ineffective when dealing with non-stationary streams or concept-drift ~\cite{lu2018learning} scenarios. On the other hand, the majority of the state-of-the-art approaches are designed to deal with image classification problems~\cite{mai2022online, qu2021recent}, but there are scarce contributions focused on other types of data, such as tabular or time series. 

Our proposal, coined as TRIL3 (Tabular-data Rehearsal Incremental LifeLong Learning method), precisely deals with these two previously mentioned aspects offering a incremental-based rehearsal method that dynamically adapts to non-stationary tabular data-streams that is highly robust to the catastrophic forgetting phenomenon. Instead of using neural networks to recreate or keep past knowledge, we apply an incremental learning model based on prototypes, XuILVQ~\cite{gonzalez2022xuilvq, gonzalez2024decentralized}, that can be trained online. XuILVQ generates previously learned examples that are replayed to feed a Deep Neural Decision Forest~\cite{kontschieder2015deep} algorithm, in charge of learning classification tasks from tabular data.

Our approach represents a step forward in CL applied to fields different from image recognition and it demonstrates strong performance on a variety of real-world datasets, effectively handling both large sample sizes and numerous features. In fact, it outperforms in most datasets other techniques designed to mitigate forgetting and the traditional offline training method.

The paper structure is the following: \Cref{sec:related} studies different continual learning approaches, mainly structured in dynamic architecture, regularization, and rehearsal, and focusing on pseudorehearsal techniques. In \Cref{sec:background} the two main algorithms used, DNDF and XuILVQ, are reviewed. Next, \Cref{sec:methodology} examines the recently introduced framework, TRIL3, explaining its working principles and workflow. \Cref{sec:validation} describes the experiments performed and the results obtained. For this purpose, the implementation details of the framework, the datasets used for the experiments, and the different tests and their results are described. \Cref{sec:conclusions} presents the main results of the paper as well as future directions, highlighting current challenges.

\section{Related work}
\label{sec:related}

In this section, we first summarize the main state-of-the-art lifelong learning techniques that have addressed one of the main limitations in lifelong learning systems: catastrophic forgetting. Ideally, the main objective in lifelong learning systems is to maintain the old knowledge (capability to perform previously acquired tasks) while new knowledge is learned. To overcome this problem, different philosophies are detailed in \Cref{sec:catastrophicforgetting}). 
After that, in \Cref{sec:rehearsalMore}), we focus on the most relevant proposals in the literature in the field of rehearsal techniques, since our methodology is precisely aligned with this idea. 

\subsection{Dealing with catastrophic forgetting in CL}
\label{sec:catastrophicforgetting}

Approaches to address the problem of catastrophic forgetting in connectionist models~\cite{french1999catastrophic, kirkpatrick2017overcoming}, such as neural networks, are usually classified according to the following three strategies: dynamic architectures~\cite{li2019learn}, regularization techniques~\cite{chen2021overcoming} and rehearsal mechanisms~\cite{robins1995catastrophic, atkinson2021pseudo}, which are detailed as follows.


\subsubsection{Dynamic architecture techniques}
\hfill\\
These techniques~\cite{li2019learn} propose to expand the neural network by adding new layers in order to have different components or subnetworks for each task that the model needs to learn. These additions can be done according to different approaches:~\cite{lesort2020continual}

\begin{itemize}
\item \textbf{Explicit Dynamic Architecture}, where any method might add, clone, or store parts of model parameters.
\item \textbf{Implicit Architecture Modification}, where the model is adapted without explicitly altering its architecture. Thus, some learning units might be deactivated or the forward-pass route might be changed. 
\item \textbf{Dual Approach}, where the neural network is divided into two models: one to learn the current task, easily adaptable, and another to keep memories of past experiences. 
\end{itemize}



\subsubsection{Regularization techniques}
\hfill\\
Regularization-based approaches~\cite{chen2021overcoming} attempt to address catastrophic forgetting by imposing constraints on the updates of neural networks. These approaches are often inspired by theoretical models from neuroscience suggesting that consolidated knowledge can be shielded from forgetting through synapses with a cascade of states producing different levels of plasticity. From a computational perspective, this is generally modeled through additional regularization terms added to the cost function that penalize changes in the mapping of a neural network during training. Regularization approaches offer a means to prevent catastrophic forgetting under certain circumstances. However, they involve additional loss terms to preserve previously acquired knowledge that  can reduce the performance of both old and new tasks when neural resources are limited, 

 In general, while they are successful in mitigating catastrophic forgetting, they have not achieved adequate results in complex contexts or datasets~\cite {mai2022online}.

\subsubsection{Rehearsal techniques}
\hfill\\
Rehearsal techniques~\cite{atkinson2021pseudo} involve replaying old data alongside the new data to prevent the model from forgetting previously learned knowledge while simultaneously learning new one. Their effectiveness usually diminishes with small buffer sizes, and they are not applicable in scenarios with data privacy concerns. It can be mainly divided into two approaches:

\begin{itemize}
\item \textbf{Pure Rehearsal}: also known as replay, raw samples are stored as a memory of past tasks at a non-negligible memory cost. 
\item \textbf{Generative Replay}, also known as \textbf{pseudo-rehearsal}, instead of using a memory buffer to store samples for replay, a generative model is trained to artificially generate data from distributions of previous tasks. A major drawback of these approaches is that generative models must be continuously updated when processing non-stationary data. 
\end{itemize}

\subsection{Rehearsal methods in the literature}
\label{sec:rehearsalMore}


Rehearsal techniques typically achieve the best performance in the current state-of-the-art~~\cite{masana2022classincremental} and they can be classified into two categories: (i) replay or buffer-based rehearsal, and (ii) generative or pseudorehearsal.

Buffer-based rehearsal typically involves storing a subset of old data samples in a memory buffer, which are then interleaved with new data during training~\cite{lopez2017gradient, rebuffi2017icarl, chaudhry2019continual}. By periodically revisiting these old samples, the model can reinforce its understanding of past tasks, thereby reducing the risk of catastrophic forgetting. This method is relatively simple and computationally efficient. However, there are two concerns while using this approach. The first is storage, specially in situations where there is not much storage space, such as in edge computing contexts or IoT settings. In these scenarios, it is sometimes not possible to store a large number of samples that are not known to what extent they will scale. The other is security, since in many cases privacy policies make it impossible to keep a memory of old data.

Pseudo-rehearsal techniques use generative models to produce synthetic data that resemble the distribution of the original training data~\cite{shin2017continual, seff2017continual, van2020brain}. Although this method does not require storing large amounts of real data, it relies heavily on the quality of the generative model to accurately capture the underlying data distribution. \Cref{pseudotable} summarizes several studies exploring different pseudo-rehearsal approaches, along with the generative models used and their respective areas of application.


\renewcommand{\arraystretch}{1.1}

\begin{table}[H]
\centering
\scriptsize
\begin{tabular}{lll}
\textbf{Publication}                                                                                                               & \multicolumn{1}{l}{\textbf{\begin{tabular}[c]{@{}l@{}}Generative \\ Model\end{tabular}}}        & \textbf{Scope} \\ \hline
\begin{tabular}[c]{@{}l@{}}Continual Learning Using World Models for  \\ Pseudo-Rehearsal~\cite{ketz2019continual} \end{tabular}                              & \multirow{5}{*}{Autoencoder}                                                                     & Image          \\
\begin{tabular}[c]{@{}l@{}}Continual Learning with Dirichlet\\  Generative-based Rehearsal~\cite{zeng2023continual} \end{tabular}                            &                                                                                                  & NLP            \\
Fearnet: Brain-inspired model for incremental learning~\cite{van2020brain}                                                                              &                                                                                                  & Image          \\
\begin{tabular}[c]{@{}l@{}}Deep generative dual memory network for continual \\ learning~\cite{kamra2017deep} \end{tabular}                              &                                                                                                  & Image          \\
\begin{tabular}[c]{@{}l@{}}Brain-inspired replay for continual\\ learning with artificial neural networks~\cite{kemker2018fearnet}  \end{tabular}             &                                                                                                  & Image          \\ \hline
\begin{tabular}[c]{@{}l@{}}Generative replay with feedback connections as a\\ general strategy for continual learning~\cite{vandeven2019generative}\end{tabular} & \multirow{4}{*}{\begin{tabular}[c]{@{}l@{}}Generative \\ Adversarial \\ Network \\ (GAN)\end{tabular}} & Image          \\
\begin{tabular}[c]{@{}l@{}}Pseudo-rehearsal: Achieving deep reinforcement learning \\ without catastrophic forgetting~\cite{atkinson2021pseudo}  \end{tabular} &                                                                                                  & Image          \\
\begin{tabular}[c]{@{}l@{}}Pseudo-Recursal: Solving the Catastrophic Forgetting \\ Problem  in Deep Neural Networks~\cite{atkinson2018pseudorecursal}\end{tabular}    &                                                                                                  & Image          \\
Continual learning with deep generative replay~\cite{shin2017continual}                                                                                       &                                                                                                  & Image          \\ \hline
\begin{tabular}[c]{@{}l@{}}Pseudorehearsal Approach for Incremental Learning\\  of Deep Convolutional Neural Networks~\cite{mellado2017pseudorehearsal}\end{tabular} & \multicolumn{1}{l}{\multirow{2}{*}{Other NN}}                                                   & Image          \\
Continual learning with invertible generative model~\cite{POMPONI2023606}                                                                                  & \multicolumn{1}{l}{}                                                                            & Image          \\ \hline
\end{tabular}
\caption{Pseudorehearsal methods evaluated.}
\label{pseudotable}
\end{table}

\Cref{pseudotable} leads us to two main conclusions. First, all studies use non-incremental generative models such as autoencoders or neural networks, which require batch training, not ideal for incremental approaches. Second, it is clear that almost all the state-of-the-art approaches were designed for image-based problems, leaving aside other kinds of domains. To understand better how these methods work, we will explore some of them in more detail.

In~\cite{atkinson2018pseudorecursal}, the authors propose a method called pseudo-recursal, which uses deep convolutional generative adversarial (DCGAN) networks to generate images representative of previous tasks. For all tasks, a 50\% ratio between real and generated data is used, except for the first task where it is 100\% real data. In addition, the model chosen as classifier is a convolutional neural network. Similarly, the model proposed in~\cite{van2020brain} integrates an image classifier and a generative model using a deep neural network architecture with pre-trained convolutional layers. The generative model, a Variational Autoencoder (VAE), facilitates generative repetition, while the classifier is based on a deep neural network with convolutional layers. Another approach, FearNet~\cite{kemker2018fearnet}, is a brain-inspired model designed for incremental class learning, used for image classification. It employs a dual memory system with three components: a recent memory system, a long-term storage system, and a decision module, which decides when information is transferred from the recent memory system to the long-term system. FearNet uses a generative autoencoder, mixing synthesized data with real data at an unspecified rate during memory consolidation.

In contrast, compared to previous work, our proposal uses an ILVQ algorithm. This incorporates an incremental generative model, which can be trained online, simultaneously with the classifier model. Moreover, the problem our approach addresses is in the scope of tabular data, which constitutes a relatively underexplored domain within the continual learning paradigm.

Additionally, only some approaches deal with non-image-related problems, like ~\cite{kurle2019continual}, where it is proposed a Bayesian neural networks for non-stationary data that trains the networks with tabular data. In~\cite{biesialska2020continual} it is conducted a survey on continual lifelong in natural language processing or in~\cite{armstrong2022continual}, where the authors evaluate different methods using time series of medical data.

\section{Background}
\label{sec:background}

In this section, we introduce the two algorithms used in our proposal (TRIL3). First, the incremental prototype-based algorithm XuILVQ~\cite{gonzalez2022xuilvq} (\Cref{sec:xuilvq}), is used to generate synthetic data to preserve old knowledge. This model generates samples from past experiences while continuously learning to map the probability distribution of unseen data over time. Second, the Deep Neural Decision Forests (DNDF) algorithm~\cite{kontschieder2015deep} (\Cref{sec:dndf}), which combines Decision Forests with Deep Neural Networks (DNN) to learn classification tasks from tabular data. 

\subsection{XuILVQ}
\label{sec:xuilvq}

XuILVQ~\cite{gonzalez2022xuilvq}, a prototype-based algorithm, is an implementation of the Incremental Learning Vector Quantization (ILVQ) algorithm~\cite{xu2012incremental} optimized for Internet of Things (IoT) scenarios. 

Prototypes are explicit representations of observations and the set of prototypes generated by the model should effectively encapsulate the essence of the entire dataset or global knowledge. XuILVQ works by incrementally updating the class prototypes according to the received data. This involves adjusting the weight vectors based on the distance between the input data and the closest prototypes. The prototypes are stored in memory, which allows the algorithm to dynamically adapt to changes in data streams without requiring historical data storage.

The algorithm works as follows: each received sample is analyzed to determine the closest prototype and the second closest prototype. If the sample either belongs to a new class or is significantly distant from both closest prototypes, it is added as a new prototype. In other cases, prototypes are updated to be grouped or ungrouped according to the class they belong to. Moreover, XuILVQ incorporates a forgetting mechanism to gradually diminish the impact of outdated data. This mechanism entails removing old isolated samples or points within low-density classes, thereby ensuring the model effectively adapts to the evolving data distribution.



This algorithm has been included in the TRIL3 framework with the aim of mapping relevant training samples and generating prototypes as adequate representations of the underlying patterns in the data over time. These prototypes are used in combination with new data during incremental model updates, avoiding catastrophic forgetting.

The main advantages of XuILVQ include its adaptive capability to dynamically adjust to changes in data patterns. Unlike other solutions, XuILVQ does not require storing old data, thus eliminating storage management issues and the need for additional resources. It offers an efficient IoT solution suitable for edge devices with limited resources. Additionally, its incremental nature facilitates online training.



\subsection{Deep Neural Decision Forests}
\label{sec:dndf}


Initially proposed by Kontschieder et al.~\cite{kontschieder2015deep}, Deep Neural Decision Forests are the other significant component of the TRIL3 framework. DNDF proposes a solution that combines decision forests with Deep Neural Networks (DNN).   

A neural decision tree is a tree structure consisting of decision and leaf nodes (also known as terminal or prediction nodes).  

Leaf nodes have an associated probability function which represents the probability distribution of the classes reaching each leaf. These functions are the first set of learnable weights that each tree uses to predict its output. 

Decision nodes are assigned a decision function that is responsible for routing samples along the tree. Every sample in a decision node is routed to the right or left subtree based on the output of the stochastic decision function.

Both probability and decision functions are differentiable, so optimal weights are obtained during model training following a Stochastic Gradient Descent (SGD) approach. 

Neural Decision Forests consist of an ensemble of neural decision trees. Predictions for every sample are obtained by averaging the output of each tree within the ensemble. 

Finally, two important properties of the DNDF model should be highlighted: Each tree within the forest is trained from a subset of input features to minimize tree correlation and counteract overfitting; Furthermore, input data are vectors encoding features from samples in batch. These vectors can be generated from a Convolution Neural Network (CNN) when applied to images or dense transformations when applied to tabular data.

Therefore, DNDF are appropriate for solving incremental learning classification tasks due to the following key aspects: 
\begin{itemize}
    \item \textbf{Problem scope:} DNDF are suitable for tabular data, which fits the scope of our scenario. Although this technique is particularly effective for tabular data, it should be noted that it can also be replaced by other differentiable models suitable for this type of data.
    \item \textbf{Performance:} DNDF outperforms conventional Decision Forests (DF) by improving their performance. By incorporating stochastic and differentiable decision functions, DNDF improves the capabilities of traditional DF.
    \item \textbf{Partially updatable:} DNDFs are inherently differentiable, which allows them to be partially updated using gradient-based optimization techniques. By taking advantage of gradient updates, DNDFs can refine their decision rules and feature representations iteratively.
\end{itemize}

\section{TRIL3: Methodology overview}
\label{sec:methodology}

TRIL3 (\textbf{T}abular-data \textbf{R}ehearsal \textbf{I}ncremental \textbf{L}ife\textbf{L}ong \textbf{L}earning) is a novel framework for continual classification of tabular data based on a pseudo-rehearsal strategy. It is mainly composed of two algorithms already detailed in \Cref{sec:background}: (i) the XuILVQ model, used to generate prototypes as suitable representations of past knowledge, (ii) a DNDF algorithm that continually learns how to perform classification tasks. Therefore, the proposed TRIL3 framework would effectively adapt to changing data distributions, facilitating continuous learning and being robust against forgetting.

\begin{figure}[H]
  \centering
    \includegraphics[width=1.08\textwidth]{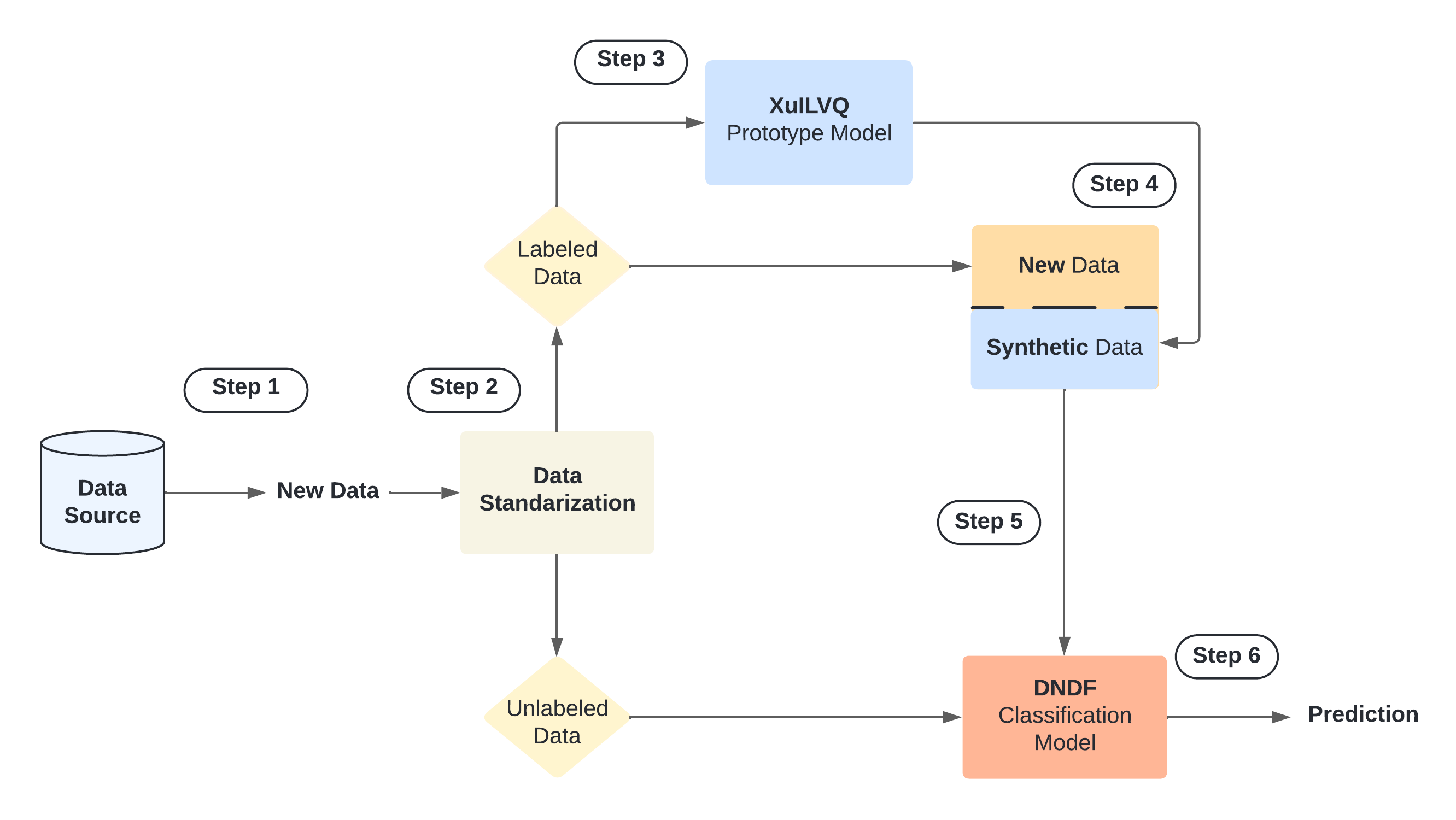}
  \caption{TRIL3 architecture and data flow}
  \label{fig:concept-architecture}
\end{figure}




\begin{algorithm}[htb]
    \caption{TRIL3}
    \label{algorithm1}
    \BlankLine
    \textbf{Input}: X: XuILVQ model ; D: DNDF model ; B : Training batch ; G : Set of prototypes used for the training batch ; S: Standardized input batch.
            \BlankLine
    \textbf{Output:} P: The prediction of the classification model.
                \BlankLine
                
    \textbf{InitializeModels()}\;
        
    \For{each input batch}{
     $S \leftarrow \textbf{Standardize}(input\ batch)$\;
      \If{input batch is labeled()}{
            $X \leftarrow \textbf{X.Update}(S)$\;
            \If{X has new prototypes}{
            $G \leftarrow \textbf{X.GeneratePrototypes()}$\;
            $B \leftarrow \textbf{GenerateTrainingBatch}(G, S)$\;
            $D \leftarrow \textbf{DF.Update}(B)$\;
        }
        }
         \If{input batch is not labeled()}{
        P = \textbf{D.MakePredictions}(S)\;
    }
    }
\end{algorithm}



 \Cref{fig:concept-architecture} depicts the proposed TRIL3 framework for online classification of tabular data, whereas the details of its behavior are summarized in the pseudo-code in \Cref{algorithm1}. The workflow is detailed in the following steps:
\begin{itemize}
    \item [-] {\bf Step 1}. In the first step, the XuILVQ and Deep Neural Decision Forests models are initialized (\textit{line 3} in \Cref{algorithm1}): the XuILVQ prototype set is initialized as empty until input data is received; on the other hand, DNDF model is initialized with random weights, setting the connection weights between neurons to values chosen randomly from a distribution before training begins.
    Small sets of data (input batches) are continuously fed into the framework, enabling online learning in real-time (\textit{line 4} in \Cref{algorithm1}). These batches may be labeled, i.e. meaning that are associated with a target value, or unlabeled. Thus, labeling provides the model with explicit information about the correct output for each input.
    
    \item [-] {\bf Step 2}. Once a batch of data has been received, it is processed as follows. First, it is online standardized (\textit{line 5} in \Cref{algorithm1}) to be consistently prepared for both the XuILVQ and DNDF models: i.e. it is adjusted to have a mean of zero and a standard deviation of one. When the data set is labeled, it is used for two purposes. On the one hand, it is used to retrain the Classification Deep Neural Decision Forests model (\textit{line 7} in \Cref{algorithm1}). On the other hand, it is used to update the XuILVQ prototype model (\textit{line 11} in \Cref{algorithm1}). Otherwise, when the input data set is an unlabeled one, it is uniquely used for prediction purposes (\textit{line 13} in \Cref{algorithm1}).
    
    \item [-] {\bf Step 3}. In the third step, the labeled batch data is used to update the prototype set using the XuILVQ algorithm. New samples are added to the prototype set if (i) there are unexpected changes in input data distributions, (ii) there are new underlying patterns or unseen classes in the batch, and (iii) the existing prototypes do not accurately represent the new samples. Otherwise, the prototype set remains unaltered. 

    \item [-] {\bf Step 4}. In Step 4 the retraining batch is formed. To this end, the distribution of samples belonging to each input batch class is calculated. For each underrepresented class within the batch, prototypes are randomly generated from the XuILVQ model to balance the input data. This enriched batch, comprised of both new data and prototypes learned from past experiences, is finally used to form the retraining batch. If new prototypes were added during the previous step (i.e., XuILVQ algorithm detects new patterns in input batch) the weights of the DNDF classification model are updated. Otherwise, no retraining is carried out and the classification model’s weights remain unaltered. This mechanism is included to avoid overfitting, in addition to ensuring that the classification model adapts to new observations without introducing unnecessary complexity when there are no significant changes in the input data.  
    
    \item [-] {\bf Step 5}. In step 5 the batch generated from the new data and synthetic data is used to train the DNDF model (\textit{line 11}  in \Cref{algorithm1}). 
    In incremental training, partial gradient updates are key to retraining an existing model from new incoming data. Partial gradient updates are carried out by performing one epoch of stochastic gradient descent on a given retraining batch. At each iteration, model parameters (i.e. both DNDF's probability and decision functions) are updated. This workflow enables the model to learn new patterns while remaining robust against forgetting. 
    
    \item [-] {\bf Step 6}. In the final step, the incoming unlabeled data is employed to make predictions using the classification model (\textit{line 13} in \Cref{algorithm1}). As part of an online process, the models are continuously updated as new data arrives.
\end{itemize}

\section{Validation and results}
\label{sec:validation}

In order to validate our proposal and compare it with other state-of-the-art approaches, we have proceeded as follows. First, we have selected representative real-world datasets related to multi-class tabular data classification problems as our data source. Then, we have designed a set of tests organized into three groups: (i) evaluate the TRIL3 performance when increasing the percentage of synthetic data to rehears the model; (ii) compare the TRIL3 performance to a state-of-the-art rehearsal-based algorithm; and (iii) compare TRIL3 performance to performance of the classical offline DNDF algorithm.

In this section, we first detail the implementation decisions in \Cref{sec:implementation}, and then we describe the selected datasets in \Cref{sec:datasets}. After that, the detailed tests and their results are described in \Cref{sec:results}, and a discussion is included in \Cref{sec:discussion}.

\subsection{Implementation details}
\label{sec:implementation}

As detailed in \Cref{sec:methodology}, TRIL3 uses two algorithms: DNDF to perform the tabular data classification tasks and XuILVQ to generate samples from past knowledge and rehearse the TRIL3 model. For the former, DNDF, we have used the implementation available at Keras~\cite{kerasdndf} as our base, although we have adapted this code to allow the model incremental retraining with one epoch of stochastic gradient descent on a given input data batch. For the latter, XuILVQ, we have used the code detailed in~\cite{gonzalez2022xuilvq}. Additionally, to perform the online standardization, we have used the StandardScaler from the library River~\cite{riverss}. All the additional code was developed in Python. \\


In order to compare the performance or TRIL3 to conventional machine learning approaches, we have used the DNDF model, a batch learning-based classification model, since the DNDF model is the one we have used to perform classification tasks in TRIL3. Instead of incrementally training the model with partial updates from incoming batches as described in \Cref{sec:methodology}), the original DNDF is fitted after processing all the training data at once. As in TRIL3, the implementation used was the one available at the Keras website~\cite{kerasdndf}, but, for this experiment, no modifications in the available code were required. \\

Finally, and with the aim of comparing our framework with other state-of-the-art lifelong learning solutions, we have used the solutions available at the Avalanche library~\cite{JMLR}.
Avalanche, developed by ContinualAI\footnote{\href{https://www.continualai.org/}{ContinualAI} is a non-profit organization for the study of CL}, is a PyTorch-based end-to-end CL library for rapid prototyping, training, and reproducible evaluation of lifelong learning algorithms. Thus, Avalanche provides several pre-built models with different strategies to mitigate catastrophic forgetting. We have selected the Replay method~\cite{bagus2021investigation}, since it also uses a rehearsal approach. The replay strategy keeps a buffer (parametrizable size) with examples of past samples that are combined with incoming data to retain old knowledge while learning new one. Additionally, Avalanche also offers some pre-built classification models suitable for tabular data. We have selected the  Multilayer Perceptron~\cite{taud2018multilayer} (MLP) model since it provides the best performance. In order to select those parameters that offer the best behavior, we have optimized the hyperparameters of the MLP and the buffer size of the Replay model, which are summarized in \Cref{Table:parameters}.

\begin{table}[H]
\centering
\scriptsize
\begin{tabular}{ccccc}
\hline
\multicolumn{5}{c}{\textbf{Parameters}}                                   \\ \hline
\multicolumn{2}{c}{MLP}     & \multicolumn{2}{c}{Optimizer} & Replay      \\ \hline
Hidden Layers & Hidden Size & Learning Rate    & Momentum   & Buffer Size \\ \hline

1             & 128         & 0.02             & 0.95       & 200         \\ \hline
\end{tabular}
\caption{Parameters used to execute the CL Replay model with a MLP neural network}
\label{Table:parameters}
\end{table}




\subsection{Datasets}
\label{sec:datasets}

In order to evaluate the performance of TRIL3, we have selected a set of representative real-world datasets related to multi-class tabular data classification problems that are summarized in \Cref{Table:datasets}. The sizes detailed in this table are the ones obtained after the datasets were processed and cleaned:

\begin{table}[H]
\centering
\scriptsize
\begin{tabular}{lccccl}
\cline{1-5}
\textbf{Dataset} & \textbf{\# Variables} & \textbf{\# Classes} & \textbf{\# Train set} & \textbf{\# Test set} &  \\ \cline{1-5}   
 CICIDS-2017 Friday   & 68                    & 2                   & 45,000                & 5,000              &  \\ \cline{1-5}
Diabetes         & 47                     & 2                   & 44,785                & 4,976                &  \\ \cline{1-5}
Wine Quality     & 11                    & 2                   & 5,847              & 650                 &  \\ \cline{1-5}
Credit Card      & 23                    & 2                   & 27,000                 & 3,000                &  \\ \cline{1-5}
\end{tabular}

\caption{Real datasets used}
\label{Table:datasets}

\end{table}

\begin{itemize}
    \item [-] \textbf{CICIDS-2017 Friday}~\cite{sharafaldin2018intrusion}, used to classify attacks, it includes detailed logs of DDoS attacks~\cite{prasad2022vmfcvd, najafimehr2022hybrid, 9031206}. Given the size of the original dataset, a subset of $50,000$ samples was selected. 
    \item [-] \textbf{Diabetes}~\cite{misc_diabetes_130-us_hospitals_for_years_1999-2008_296} is used to predict patient readmission~\cite{kuriakose2022prediction, tamin2017implementation} and contains medical records from $130$ hospitals in the US, spanning the years 1999 to 2008. We used only those features that are integers.
    \item [-] \textbf{Wine Quality}~\cite{misc_wine_quality_186} includes chemical properties of the wine and is used to predict quality indexes or, as in this case, the color of the wine~\cite{gupta2018selection, dahal2021prediction}. 
    \item [-] \textbf{Credit Card}~\cite{misc_default_of_credit_card_clients_350} contains information about credit card customers with the aim of predicting default payment next month~\cite{mqadi2021solving, alam2020investigation}.

\end{itemize}

\subsection{Tests and results}
\label{sec:results}

As it was aforementioned, our purpose is to validate the TRIL3 approach by comparing its performance (parameter $F1$) to other alternatives in the state-of-the-art and also to check its behavior when different percentages of synthetic data are used for rehearsal. All the obtained results are summarized in \Cref{tab:F1Scores}. In this table:

\begin{itemize}
    \item [-] Column TRIL3 groups the $F1$ score obtained using different radios between real and synthetic samples within each incoming training batch. This parameter is taken into account in step 4 of the TRIL3 workflow (\Cref{fig:concept-architecture}). Please, take into account that $100\%$ percentage of real data within the training batch corresponds to a naive execution, that is an incremental execution without any mechanisms against catastrophic forgetfulness.
    \item [-] Column Replay in \Cref{tab:F1Scores} groups the performance results in a state-of-the-art rehearsal-based algorithm: a multi-layer perceptron neural network with a Replay strategy, described in \Cref{sec:implementation}.
    \item [-] Finally, column Offline DNDF  groups the performance results obtained using the offline learning version of DNDF also described in \Cref{sec:implementation}. 
\end{itemize}

\begin{table}[H]
\centering
\footnotesize
\begin{tabular}{cclllccccc}
\hline
\multicolumn{8}{c|}{TRIL3}                                                                                                                                                                                     & \multicolumn{1}{c|}{Replay}        & \begin{tabular}[c]{@{}c@{}}Offline\\ DNDF\end{tabular} \\ \hline
\multicolumn{5}{c|}{\textbf{Real Data Ratio (\%) }}                                                    & \multicolumn{1}{c|}{\textbf{50}}   & \multicolumn{1}{c|}{\textbf{75}}   & \multicolumn{1}{c|}{\textbf{100}}  & \multicolumn{1}{c|}{-}             & -                                                      \\ \hline \hline

\multicolumn{10}{c}{CICIDS-2017 Friday}                                                                                                                                                                                                                                                                                          \\ \hline
\multicolumn{1}{c|}{\multirow{2}{*}{Class 0}} & \multicolumn{4}{c|}{F1-score Before Forgetting Phase} & \multicolumn{1}{c|}{\textbf{0.98}} & \multicolumn{1}{c|}{0.97}          & \multicolumn{1}{c|}{0.96}          & \multicolumn{1}{c|}{\textbf{0.98}} & \multirow{2}{*}{0.99}                                  \\
\multicolumn{1}{c|}{}                         & \multicolumn{4}{c|}{F1-score During Forgetting Phase} & \multicolumn{1}{c|}{\textbf{0.99}} & \multicolumn{1}{c|}{\textbf{0.99}} & \multicolumn{1}{c|}{0.8}           & \multicolumn{1}{c|}{\textbf{0.99}} &                                                        \\ \cline{1-5}
\multicolumn{1}{c|}{\multirow{2}{*}{Class 1}} & \multicolumn{4}{c|}{F1-score Before Forgetting Phase} & \multicolumn{1}{c|}{\textbf{0.98}} & \multicolumn{1}{c|}{0.97}          & \multicolumn{1}{c|}{\textbf{0.98}} & \multicolumn{1}{c|}{\textbf{0.98}} & \multirow{2}{*}{0.99}                                  \\
\multicolumn{1}{c|}{}                         & \multicolumn{4}{c|}{F1-score During Forgetting Phase} & \multicolumn{1}{c|}{\textbf{0.99}} & \multicolumn{1}{c|}{\textbf{0.99}} & \multicolumn{1}{c|}{0.74}          & \multicolumn{1}{c|}{\textbf{0.99}} &                                                        \\ \hline \hline

\multicolumn{10}{c}{Diabetes}                                                                                                                                                                                                                                                                                      \\ \hline
\multicolumn{1}{c|}{\multirow{2}{*}{Class 0}} & \multicolumn{4}{c|}{F1-score Before Forgetting Phase} & \multicolumn{1}{c|}{\textbf{0.48}} & \multicolumn{1}{c|}{0.32}          & \multicolumn{1}{c|}{0}             & \multicolumn{1}{c|}{0.33}          & \multirow{2}{*}{0.21}                                  \\
\multicolumn{1}{c|}{}                         & \multicolumn{4}{c|}{F1-score During Forgetting Phase} & \multicolumn{1}{c|}{\textbf{0.52}} & \multicolumn{1}{c|}{0.51}          & \multicolumn{1}{c|}{0.46}          & \multicolumn{1}{c|}{0.45}          &                                                        \\ \cline{1-5}
\multicolumn{1}{c|}{\multirow{2}{*}{Class 1}} & \multicolumn{4}{c|}{F1-score Before Forgetting Phase} & \multicolumn{1}{c|}{0.65}          & \multicolumn{1}{c|}{0.73}          & \multicolumn{1}{c|}{\textbf{0.8}}  & \multicolumn{1}{c|}{0.74}          & \multirow{2}{*}{0.8}                                   \\
\multicolumn{1}{c|}{}                         & \multicolumn{4}{c|}{F1-score During Forgetting Phase} & \multicolumn{1}{c|}{0.60}          & \multicolumn{1}{c|}{0.19}          & \multicolumn{1}{c|}{0.08}          & \multicolumn{1}{c|}{\textbf{0.64}} &                                                        \\ \hline \hline
\multicolumn{10}{c}{Credit Card}                                                                                                                                                                                                                                                                                   \\ \hline
\multicolumn{1}{c|}{\multirow{2}{*}{Class 0}} & \multicolumn{4}{c|}{F1-score Before Forgetting Phase} & \multicolumn{1}{c|}{0.83}          & \multicolumn{1}{c|}{\textbf{0.88}} & \multicolumn{1}{c|}{\textbf{0.88}} & \multicolumn{1}{c|}{0.87}          & \multirow{2}{*}{0.89}                                  \\
\multicolumn{1}{c|}{}                         & \multicolumn{4}{c|}{F1-score During Forgetting Phase} & \multicolumn{1}{c|}{0.88}          & \multicolumn{1}{c|}{\textbf{0.9}}  & \multicolumn{1}{c|}{0.87}          & \multicolumn{1}{c|}{0.84}          &                                                        \\ \cline{1-5}
\multicolumn{1}{c|}{\multirow{2}{*}{Class 1}} & \multicolumn{4}{c|}{F1-score Before Forgetting Phase} & \multicolumn{1}{c|}{\textbf{0.51}} & \multicolumn{1}{c|}{0.43}          & \multicolumn{1}{c|}{0.24}          & \multicolumn{1}{c|}{0.34}          & \multirow{2}{*}{0.43}                                  \\
\multicolumn{1}{c|}{}                         & \multicolumn{4}{c|}{F1-score During Forgetting Phase} & \multicolumn{1}{c|}{\textbf{0.55}} & \multicolumn{1}{c|}{0.5}           & \multicolumn{1}{c|}{0.03}          & \multicolumn{1}{c|}{0.45}          &                                                        \\ \hline \hline

\multicolumn{10}{c}{Wine}                                                                                                                                                                                                                                                                                          \\ \hline
\multicolumn{1}{c|}{\multirow{2}{*}{Class 0}} & \multicolumn{4}{c|}{F1-score Before Forgetting Phase} & \multicolumn{1}{c|}{0.88}          & \multicolumn{1}{c|}{0.67}          & \multicolumn{1}{c|}{0.35}          & \multicolumn{1}{c|}{\textbf{0.96}} & \multirow{2}{*}{0.99}                                  \\
\multicolumn{1}{c|}{}                         & \multicolumn{4}{c|}{F1-score During Forgetting Phase} & \multicolumn{1}{c|}{0.94}          & \multicolumn{1}{c|}{0.97}          & \multicolumn{1}{c|}{0.89}          & \multicolumn{1}{c|}{\textbf{0.99}} &                                                        \\ \cline{1-5}
\multicolumn{1}{c|}{\multirow{2}{*}{Class 1}} & \multicolumn{4}{c|}{F1-score Before Forgetting Phase} & \multicolumn{1}{c|}{0.95}          & \multicolumn{1}{c|}{0.95}          & \multicolumn{1}{c|}{0.88}          & \multicolumn{1}{c|}{\textbf{0.97}} & \multirow{2}{*}{0.99}                                  \\
\multicolumn{1}{c|}{}                         & \multicolumn{4}{c|}{F1-score During Forgetting Phase} & \multicolumn{1}{c|}{\textbf{0.99}} & \multicolumn{1}{c|}{\textbf{0.99}} & \multicolumn{1}{c|}{0.97}          & \multicolumn{1}{c|}{\textbf{0.99}} &                                                        \\ \hline

\end{tabular}
\caption{Performance comparison between TRIL3 and (i) an online rehearsal-based solution (Replay) and (ii) an offline DNDF solution. The table also includes the performance variation of TRIL3 according to different percentages of synthetic data used for rehearsal.}
\label{tab:F1Scores}
\end{table}

The tests were carried out as follows. First, the batches of data, which include samples of all existing classes, are processed by the learning model. This entails the model learning about all the classes, i.e. the complete problem. After that, and during a long period called \textit{forgetting phase}, class 1 is deliberately omitted from the input data. In a conventional system, this absence would lead to the degradation of the model concerning this particular class that, eventually, would be erased from global knowledge. This happens because the model tends to overfit the remaining classes during the continuous retraining. In a CL model, this problem should be avoided by using the appropriate techniques to avoid catastrophic forgetting. 




\begin{figure}[H]
  \centering
  \includegraphics[width=135mm,scale=1]{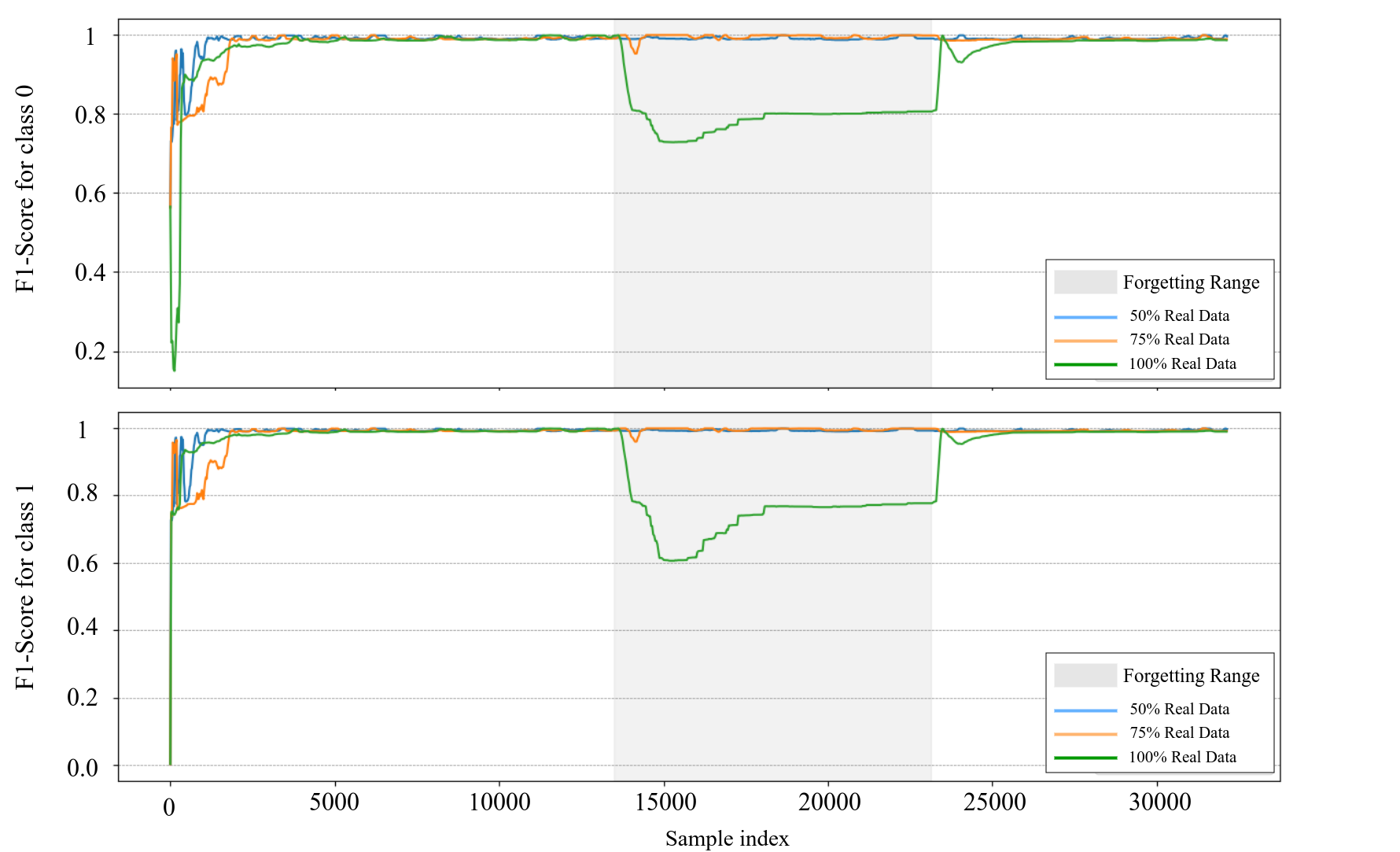}
  \caption{F1-Score class 0 and 1, CICIDS-2017 Friday dataset}
  \label{fig:graph1_ddos}
\end{figure}

This evolution is depicted in \Cref{fig:graph1_ddos} for the CICIDS-2017 Friday dataset, the most popular. Each graph shows the F1-score for individual classes on the y-axis, with the input batch number delineated along the x-axis. The gray area represents the forgetting phase, where class 1 disappears from the input batches. It must be highlighted that the performance of TRIL3 with $25\%$ and $50\%$ of synthetic data within each retraining batch preserves the model from forgetting. However, when the retraining batch is $100\%$ formed by real data (i.e. the naive solution, as DNDF has no mechanisms to resist forgetting) the performance of the model drops during the forgetting phase, as expected. 

\begin{figure}[H]
  \centering
    \includegraphics[width=1.08\textwidth]{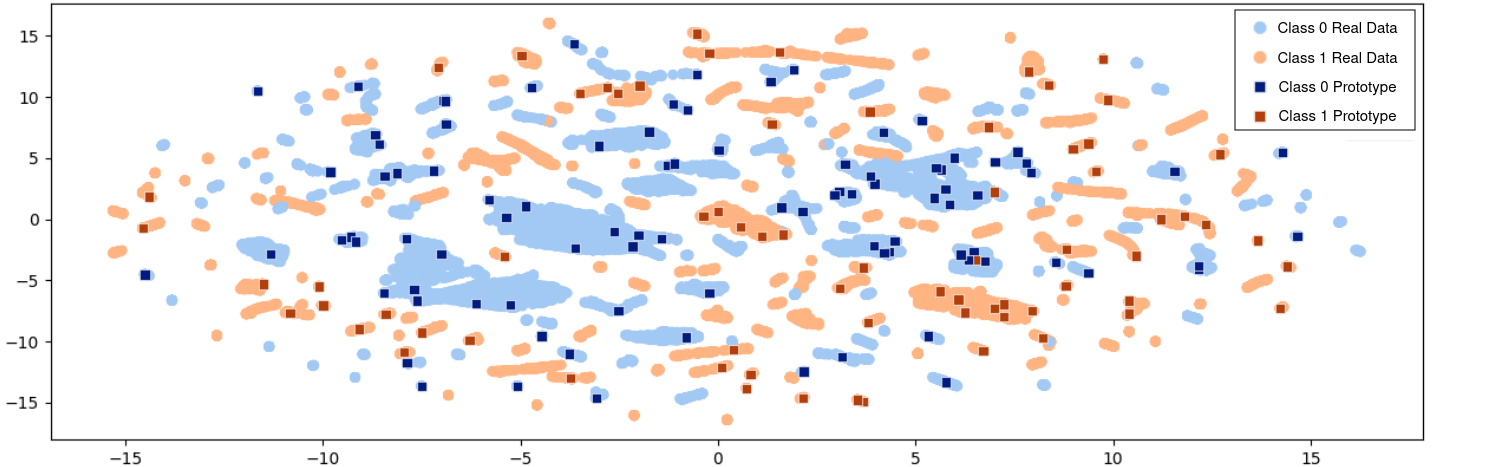}
  \caption{Real data and prototypes, CICIDS-2017 Friday dataset}
  \label{fig:prototypes}
\end{figure}

Additionally, \Cref{fig:prototypes} shows a scatter plot generated by UMAP~\cite{mcinnes2020umap}, a dimension reduction technique, that allows to look at samples by similarity in a low-dimensional space. In the image, two classes of data are represented: class $0$, in blue, and class $1$, in red. Furthermore, the circular dots indicate the real samples of each class, while the squares represent the prototypes. As can be seen, the prototypes, with a much reduced number of points, represent the distribution of the real data very accurately.

During the course of the experiment, the number of prototypes changes as they are dynamically updated and deleted. The image captures the state of these prototypes at the final moment of the experiment. In this dataset for class $0$, the peak number of prototypes never exceeded $100$, while for class $1$ the maximum number of prototypes remained under $80$.



\subsection{Discussion}
\label{sec:discussion}

As \Cref{tab:F1Scores} shows, TRIL3 with $50\%$ synthetic data consistently shows outstanding results in all analyzed datasets. For example, the CICIDS-2017 Friday dataset achieves an almost perfect F1-score ($0.98$-$0.99$) both before and during the forgetting phase for both classes. Moreover, in the diabetes dataset, it stands out especially for class 0 before the forgetting phase ($0.48$) and improves performance during this phase ($0.52$). In the credit card dataset, TRIL3 with $75\%$ and $50\%$ synthetic data shows the best results for classes $0$ and $1$ respectively, evidencing its ability to handle different data distributions effectively. Finally, in the wine dataset, TRIL3 with $50\%$ real data shows exceptional performance both before and during the forgetting phase. On the other hand, the model performance notably drops with $100\%$ of real data within the training batch, that is, when the DNDF classification model has no support to retain past knowledge during the forgetting phase.

Compared to the MLP method with replay implemented using Avalanche, TRIL3 shows several advantages in the analyzed datasets. In the case of the CICIDS-2017 Friday dataset, there was a tie between TRIL3 with $50\%$ synthetic data, replay, and offline training, achieving in all cases a near-perfect F1 score for both classes. On the diabetes and credit card datasets, TRIL3 results are generally better than Replay. Finally, in the wine dataset, the smallest one, MLP with replay performs better. Therefore, in comparison with the MLP with the replay method, TRIL3 obtains equal or better results in all the datasets except for the wine dataset.

In terms of performance compared to traditional offline training, TRIL3 also shows promising results. In all studied datasets, TRIL3 manages to approach or improve the performance of batch offline training. On Friday CICIDS-2017 dataset, both achieve equal maximum performance. In diabetes and credit card datasets, where offline performance drops for certain classes, TRIL3 demonstrates better capability except for class 0 of the diabetes dataset. Finally, in the wine dataset, TRIL3 performs worse than offline training.

Overall, TRIL3 with $50\%$ synthetic data is the best performer in most datasets, matching or even outperforming batch training, followed by MLP with replay. In the case of naive online training, as expected, there is a generalized drop in performance during the forgetting phase.

\section{Conclusions}
\label{sec:conclusions}

In this paper, we introduce a new methodology, coined as TRIL3. This novel continual learning system is designed to address the problem of catastrophic forgetting in the online classification of tabular data. Our approach integrates the prototype-based incremental generative model XuILVQ to generate synthetic data to preserve old knowledge and the DNDF algorithm, which was modified to run in an incremental way, to learn classification tasks for tabular data. This proposal, based on a rehearsal strategy, does not require the storage of old samples and it is able to adapt to non-stationary data streams, being robust against catastrophic forgetting.

We have assessed TRIL3 comparing its performance to other solutions in the literature using four representative real-world datasets, available online, for multi-class tabular data classification problems. Using this data, we have compared TRIL3 with another CL approach based also on a rehearsal philosophy (Replay) that is available at the Avalanche library and we have also compared TRIL3 with the classical DNDF model. In both cases, the performance of TRIL3 generally outstands the other options. Besides, and after analyzing which percentage of synthetic data would be the most adequate to rehearse the model, the results indicate that $50\%$ is the best option. Thus, TRIL3 not only offers good results according to the system performance but it also offers a good solution against catastrophic forgetting.


In future work, we plan to extend our framework to address classification problems involving multiclass target variables. Although promising results have been obtained using synthetic data sets, it is crucial to identify high-quality real data sets for further validation. In addition, exploring the adaptation of this framework for use with time series data is an interesting possibility. This adaptation could offer advantages for industrial applications.

\section{Acknowledgements}
This work was supported by the Centro Tecnolóxico de Telecomunicacións de Galicia (GRADIANT) and by the grant PID2020-113795RB-C33 funded by MICIU/AEI/10.13039/501100011033 (COMPROMISE project). Additionally, this work has also received financial support from Consellería de Educación, Ciencia, Universidades e Formación Profesional and co-funded by UE.

\bibliographystyle{ieeetr}


\end{document}